\documentclass[times, review, 10pt]{elsarticle}

\usepackage{amsmath,amsfonts}
\usepackage{algorithmic}
\usepackage{algorithm}
\usepackage{array}
\usepackage[caption=false,font=normalsize,labelfont=sf,textfont=sf]{subfig}
\usepackage{textcomp}
\usepackage{stfloats}
\usepackage{url}
\usepackage{verbatim}
\usepackage{graphicx}
\usepackage{alltt}
\usepackage{booktabs}
\usepackage{multirow}
\usepackage{varwidth}
\usepackage{comment}
\usepackage{xcolor}
\definecolor{customyellow}{RGB}{241, 248, 249}
\definecolor{customred}{RGB}{183, 219, 227}
\usepackage{bbding}
\usepackage{colortbl} 
\usepackage{caption}
\usepackage{subcaption} 
\usepackage{makecell}
\usepackage{enumitem}
\usepackage{newfloat}
\usepackage{listings}
\usepackage{subcaption}
\usepackage{float}
\usepackage{bm}
\usepackage{amssymb}
\usepackage{capt-of}
\usepackage{caption}
\usepackage{tabularx}
\usepackage{lipsum}  
\usepackage{wrapfig}
\usepackage{graphicx}
\usepackage{pifont}

\definecolor{MidnightBlue}{rgb}{0.1, 0.1, 0.44}
\definecolor{Bittersweet}{rgb}{1.0, 0.44, 0.37}

\journal{Journal of \LaTeX\ Templates}

\begin{document}

\begin{frontmatter}

\title{Class-Aware Prototype Learning with Negative Contrast for Test-Time Adaptation of Vision-Language Models} 

\author[1,2]{Xiaozhen Qiao}
\author[2]{Jingkai Zhao}
\author[2]{Yuqiu Jiang}
\author[3]{Xianda Guo}
\author[2,4]{Zhe Sun\corref{correspond}}
\author[2]{Hongyuan Zhang\corref{correspond}}
\author[2]{Xuelong Li\corref{correspond}}

\address[1]{School of Information Science and Technology, University of Science and Technology of China, 100 Fuxing Street, Hefei 230026, P. R. China }
\address[2]{Institute of Artificial Intelligence (TeleAI), China Telecom, P. R. China.}
\address[3]{College of Computer Science, Wuhan University, Wuhan 430072, P. R. China}

\address[4]{School of Artificial Intelligence, OPtics and ElectroNics (iOPEN), 
Northwestern Polytechnical University, Xi'an 710072, P. R. China}

\cortext[correspond]{Corresponding author}

\begin{abstract}

Vision-Language Models (VLMs) demonstrate impressive zero-shot generalization through large-scale image-text pretraining, yet their performance can drop once the deployment distribution diverges from the training distribution. To address this, Test-Time Adaptation (TTA) methods update models using unlabeled target data. However, existing approaches often ignore two key challenges: prototype degradation in long-tailed distributions and confusion between semantically similar classes. To tackle these issues, we propose \textbf{C}lass-Aware \textbf{P}rototype \textbf{L}earning with \textbf{N}egative \textbf{C}ontrast(\textbf{CPL-NC}), a lightweight TTA framework designed specifically for VLMs to enhance generalization under distribution shifts. CPL-NC introduces a \textit{Class-Aware Prototype Cache} Module that dynamically adjusts per-class capacity based on test-time frequency and activation history, with a rejuvenation mechanism for inactive classes to retain rare-category knowledge. Additionally, a \textit{Negative Contrastive Learning} Mechanism identifies and constrains hard visual-textual negatives to improve class separability. The framework employs asymmetric optimization, refining only textual prototypes while anchoring on stable visual features. Experiments on 15 benchmarks show that CPL-NC consistently outperforms prior TTA methods across both ResNet-50 and ViT-B/16 backbones.

\end{abstract}










\end{frontmatter}

\section{Introduction}

Large-scale Vision-Language Models (VLMs), such as CLIP~\cite{radford2021learning}, ALIGN~\cite{jia2021scaling} and SigLIP~\cite{zhai2023sigmoid}, have demonstrated remarkable zero-shot generalization by learning a unified multimodal semantic space through contrastive pre-training on large-scale image-text pairs. These models enable recognition of unseen categories without additional training and have become a cornerstone of modern visual understanding~\cite{tao2008bayesian} across tasks such as image classification~\cite{karmanov2024efficient, zhang2024dual,huang2025enhance,jiang2025mixture}, object detection~\cite{kim2024vlm,madan2024revisiting,gu2025contrastive}, object tracking~\cite{sun2024chattracker, shao2024context}, and cross-modal retrieval~\cite{li2025test}. However, their generalization ability remains limited under domain shifts and out-of-distribution conditions~\cite{heng2025detecting}. Test-time distributions are frequently subject to substantial shifts due to background clutter, visual heterogeneity, pose variability, and class imbalance~\cite{tu2024empirical}. These factors compromise the model’s semantic alignment and discriminative power, leading to prediction uncertainty and semantic drift. The divergence between pre-training and test-time distributions substantially hinders the generalization of VLMs~\cite{li2022positive}, especially under cross-domain and out-of-distribution conditions.

To reduce the impact of distribution shift, recent studies~\cite{shu2022test, feng2023diverse,ma2024retta,ye2025domain,tian2024test} have explored Test-Time Adaptation (TTA) techniques that update models during testing without using source data or target labels. These methods improve model performance on shifted data by making small updates for each test sample. Test-Time Prompt Tuning (TPT)~\cite{shu2022test} improves test-time performance by adjusting prompts using multiple augmented views and selecting confident predictions. DiffTPT\cite{feng2023diverse} extended this idea with diffusion-based augmentations to improve visual diversity and adaptation robustness. More recently, DPE~\cite{zhang2024dual} improves both visual and text prototypes during testing, helping the model better align images and texts. These methods have extended TTA from tuning prompts to updating class prototypes, making vision-language models more adaptable without using any labels.

\begin{figure*}[t]
  \centering
  \includegraphics[width=1.00\textwidth]{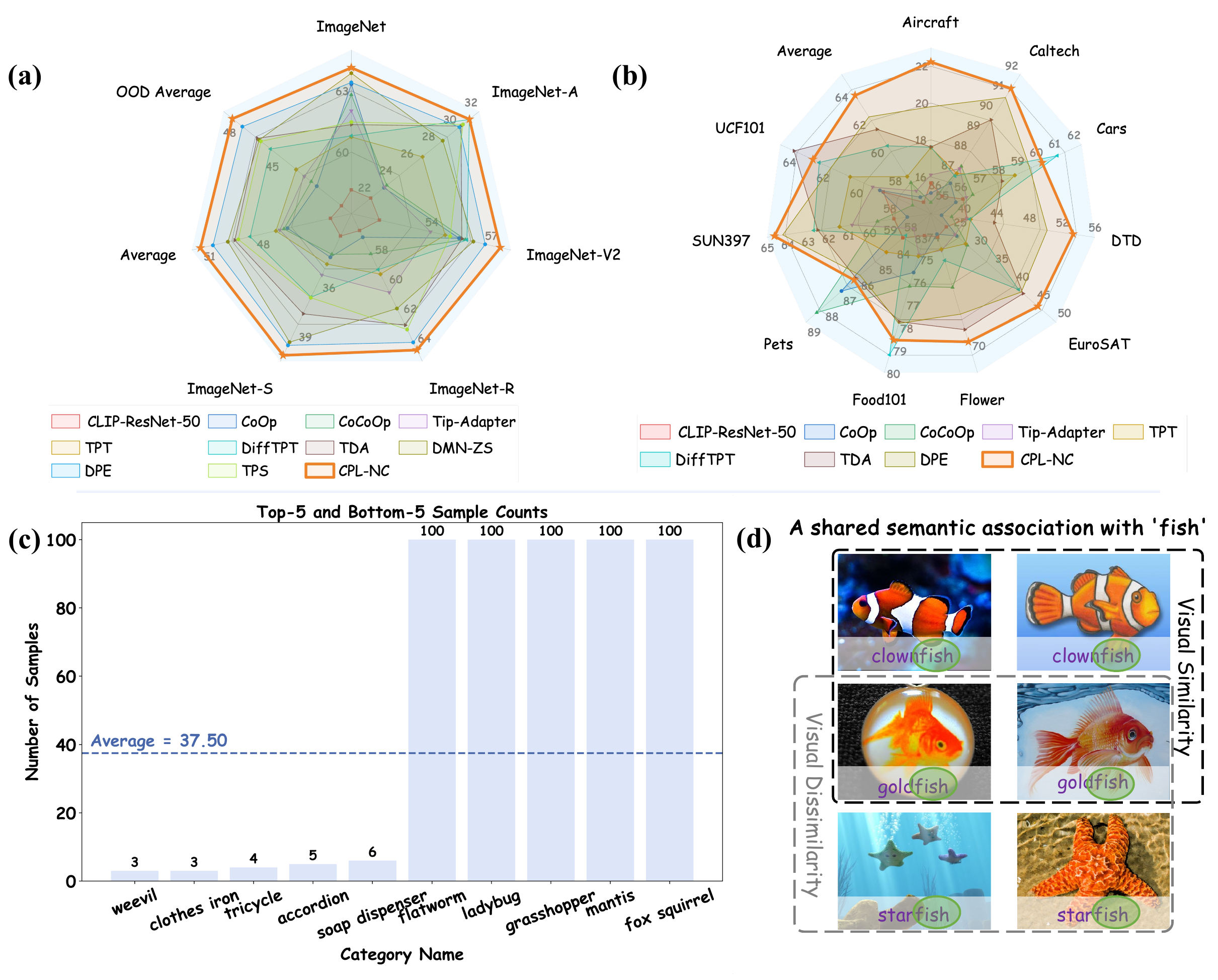}
  \caption{Comparative results of \textit{CPL-NC} on (a) OOD robustness using ViT-B/16 and (b) cross-dataset generalization using ResNet-50.(c) Sample distribution of the top-5 and bottom-5 categories in ImageNet-A~\cite{hendrycks2021natural}, illustrating significant class imbalance. Some minority classes account for less than 1\% of total samples, while dominant classes far exceed the average (dashed line). (d) Visual heterogeneity under the shared semantic label "fish" in ImageNet-R~\cite{hendrycks2021many}. While clownfish and goldfish are visually similar, starfish differs substantially in appearance despite sharing the same semantic root. }
  \label{fig:1}
  \vspace{-20pt}
\end{figure*}
Despite their effectiveness, existing methods still face two critical problems that limit their performance under complex distribution shifts. First, they neglect class imbalance and long-tail activation patterns. Most approaches (e.g., TDE~\cite{karmanov2024efficient}, DPE~\cite{zhang2024dual}) adopt a fixed cache size for all classes, ignoring frequency disparities or dynamic emergence of classes. As shown in Figure~\ref{fig:1}(c), benchmarks such as ImageNet-A~\cite{hendrycks2021natural} exhibit extreme imbalance, with some tail categories comprising less than 1\% of samples. During test time, some infrequent classes may occur sparsely, sometimes appearing only in the early phases of adaptation. Their prototypes, affected by early noise, can easily be weakened or removed when more common classes are updated. We call this problem '\textit{dead classes}'. Fixed-capacity caches cannot prevent this, leading to prototype forgetting and reduced robustness for tail categories. Second, these methods lack mechanisms to disambiguate semantically similar classes. Most methods rely purely on inner-product similarity between image and text embeddings, they fail to explicitly enforce decision boundaries between confusing categories. Figure~\ref{fig:1}(d) shows that semantically related classes (e.g., “goldfish” vs. “starfish”) may differ greatly in appearance, causing confusion in the absence of explicit disambiguation—especially in fine-grained or open-vocabulary tasks.

To address these issues, we propose CPL-NC \textit{(Class-Aware Prototype Learning with Negative Contrast)}, a test-time adaptation framework tailored for VLMs, as illustrated in Figure~\ref{fig:pipeline}. CPL-NC introduces two key modules targeting structural memory and semantic discrimination. First, the \textit{Class-Aware Prototype Cache} Module dynamically adjusts per-class cache capacities based on both test-time frequency and recency. A nonlinear suppression function redistributes capacity to favor rare classes, enhancing fault tolerance and representation fidelity. In addition, we propose a Dead-Class Rejuvenation mechanism that compensates long-inactive classes with temporary capacity boosts and synthetic features from visually or semantically similar categories, preserving their prototype viability under long-tailed distributions. Second, the \textit{Negative Contrastive Learning} Mechanism selects the most similar but incorrect visual-textual prototype pairs and applies an InfoNCE loss to explicitly separate confusing classes in the embedding space. This mechanism enhances discriminability near fine-grained decision boundaries and reduces ambiguity between semantically similar categories. Combined with entropy minimization and modality alignment losses, it forms a unified optimization objective that improves both confidence and consistency. To reduce computational overhead and improve stability, CPL-NC employs an asymmetric refinement strategy: cached visual prototypes act as relatively stable anchors but are still incrementally refreshed, while textual prototypes undergo parametric refinement with higher update flexibility. This design enables efficient adaptation while preserving accumulated knowledge.

We evaluate CPL-NC across 15 image recognition benchmarks exhibiting natural domain shifts. As shown in Figure~\ref{fig:1}~(a,b), CPL-NC consistently outperforms existing TTA methods across diverse tasks and architectural backbones. In summary, our main contributions are as follows:
\begin{itemize}
    \item We propose a \textbf{Class-Aware Prototype Cache} module that dynamically allocates cache capacity based on test-time frequency and activation history, with a rejuvenation strategy to preserve long-tail representations.
    \item We introduce a \textbf{Negative Contrastive Learning} mechanism that mines hard visual-textual negatives to enhance fine-grained class separation, leveraging an \textit{asymmetric optimization} strategy that updates to lightweight textual parameters for stability and efficiency.
    \item Extensive experiments on 15 cross-domain and open-world benchmarks validate the superior adaptability and robustness of the proposed CPL-NC framework over existing test-time adaptation methods.
\end{itemize}

\section{Realted Work}

\subsection{Vision-Language Models}
Vision-Language Models (VLMs) like CLIP~\cite{radford2021learning} , ALIGN~\cite{jia2021scaling} and SigLIP~\cite{zhai2023sigmoid} have demonstrated remarkable success by leveraging large-scale image-text pre-training, enabling zero-shot transfer to tasks such as image classification~\cite{hegde2023clip} and object detection~\cite{wu2023cora}. To further improve downstream adaptability, various supervised finetuning paradigms have been proposed. Prompt-based methods (e.g., CoOp~\cite{zhou2022learning}, CoCoOp~\cite{zhou2022conditional}) learn task-specific textual prompts to better align with image embeddings. Adapter-based approaches (e.g., Tip-Adapter~\cite{zhang2022tip}, TaskRes~\cite{yu2023task}) introduce lightweight modules that can be trained on limited labeled data to adapt VLMs to new domains~\cite{gao2024clip, zhou2024dynamic}. While effective, these methods depend on labeled target data and thus are inapplicable in test-time adaptation (TTA) scenarios, where no labels are available at inference. In contrast, CPL-NC enables test-time adaptation without labeled target data. By integrating class-aware caching and negative contrastive learning, CPL-NC refines cross-modal prototypes during inference, making it highly effective in real-world scenarios where labeled data is impractical or unavailable. This method not only addresses issues of prototype degradation and class imbalance but also improves the model's robustness and generalization.

\subsection{Test-Time Adaptation}
Test-time adaptation addresses domain shift by enabling dynamic model updates using unlabeled test samples~\cite{tang2023neuro, karmanov2024efficient, zhang2024dual}. This approach has become essential for maintaining model performance when exposed to out-of-distribution data. TPT~\cite{shu2022test} was one of the first to explore this, using consistency across augmented views of test samples to improve model robustness. Following this, DiffTPT~\cite{feng2023diverse} introduced diffusion-based augmentations, and C-TPT~\cite{yoon2024c} focused on mitigating calibration errors. Memory-based methods like TDA~\cite{karmanov2024efficient} and DMN~\cite{zhang2024dual0} further advanced test-time adaptation by enabling adaptation across multiple test samples, marking a shift from single-sample to collective adaptation strategies.
DPE~\cite{zhang2024dual} enhanced this by introducing dual-modality prototype evolution, dynamically updating prototypes during testing to improve feature modeling. However, while these methods focus on prototype evolution, they overlook the interplay between prototype quality and sample quality during adaptation.
CPL-NC, on the other hand, addresses this gap by combining class-aware prototype evolution with sample quality. By dynamically refining prototypes during inference, CPL-NC not only enhances adaptation without labeled data but also improves generalization to unseen data. With integrated class-aware cache resizing and negative contrastive learning, CPL-NC provides a robust solution to the challenges of class imbalance, prototype forgetting, and semantic ambiguity in test-time adaptation, outperforming previous methods.

\begin{figure*}[!t]
 \centering
\includegraphics[width=1.0\linewidth]{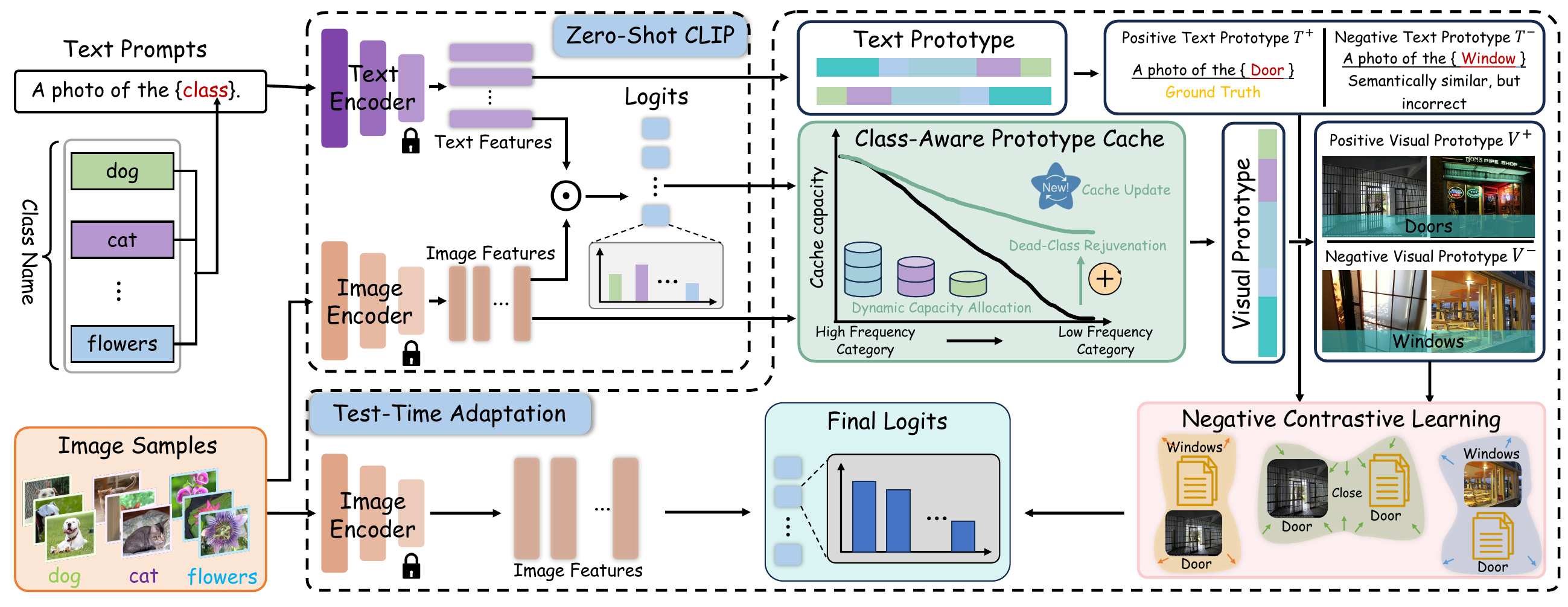}
\caption{\textbf{Overview of the CPL-NC framework.} Given an input image $X$, CPL-NC extracts visual features $f_v$ and retrieves visual prototypes $v_c$ from a class-aware cache $\mathcal{C}$. Textual prototypes $t_c$ are refined to $\hat{t}_c$ during test time. The cache dynamically adjusts class capacity $M_c$ based on frequency $p_c$ and inactivity, applying a decay-aware boost $M_c^{\text{boost}}$ for rare classes. Final predictions combine $f_v$, $v_c$, and $\hat{t}_c$, with cache updates triggered by low entropy. A negative contrastive module selects hard negatives $(v_c^{-}, t_c^{-})$ to enforce InfoNCE loss $\mathcal{L}_{\text{NCL}}$, improving semantic separation by updating only textual prototypes.}
\label{fig:pipeline}  
\vspace{-10pt}
\end{figure*}

\section{Methods}

As illustrated in Figure~\ref{fig:pipeline}, we introduce CPL-NC (Class-Aware Prototype Learning with Negative Contrast), a test-time adaptation framework for VLMs. CPL-NC performs lightweight test-time updates without source data, combining class-aware caching, rare class activation, and negative contrastive learning to refine cross-modal prototypes. It addresses class imbalance, prototype degradation, and semantic ambiguity to improve robustness and generalization. 

\subsection{Preliminary}

\noindent \textbf{CLIP Overview.} CLIP~\cite{radford2021learning} is comprised of two distinct encoders: a visual encoder \(\mathcal{E}_v(\cdot)\) and a textual encoder \(\mathcal{E}_t(\cdot)\). These encoders independently project images and class-level textual descriptions into a shared \(d\)-dimensional embedding space, thereby facilitating cross-modal alignment. For zero-shot classification over \(C\) categories, the prediction for a test image \(X_{\mathtt{test}} \in \mathcal{D}_{\mathtt{test}}\) is obtained by computing the similarity between the image embedding and the embeddings of class-specific text descriptions \(\mathcal{T}_c\):
\begin{align}
& f_v = \mathcal{E}_v(X_{\mathtt{test}}), \quad f_{t_c} = \mathcal{E}_t(\mathcal{T}_c) \nonumber \\
& \mathbb{P}_{\text{CLIP}}(y = y_c | X_{\mathtt{test}}) = \frac{\exp(\mathrm{sim}(f_{t_c}, f_v)/\tau)}{\sum\limits_{c'} \exp(\mathrm{sim}(f_{t_{c'}}, f_v)/\tau)},
\end{align}
where \(\mathrm{sim}(\cdot, \cdot)\) denotes cosine similarity, and \(\tau\) is a learnable temperature parameter. This formulation enables CLIP to perform zero-shot recognition by leveraging semantic information encoded in textual prompts.



\noindent \textbf{Test-Time Prompt Adaptation.} To enhance the robustness of CLIP under domain shift, Test-Time Prompt Tuning (TPT)~\cite{shu2022test} has been introduced. In this approach, prompts are dynamically adapted using augmented views of test samples. Given $N$ augmentations $\{\mathcal{A}_n(X_{\mathtt{test}})\}_{n=1}^N$, predictions with entropy below a threshold $t$ are selected, and the final output is computed by averaging over the top $\rho$-percentile based on prediction confidence:

\begin{equation}
\mathbb{P}_{\mathtt{TPT}}(X_{\mathtt{test}}) = \frac{1}{\rho N} \sum_{n=1}^{N} \mathbf{1}\Big[\mathcal{H}\big(\mathbb{P}_{\mathtt{CLIP}}(\mathcal{A}_n(X_{\mathtt{test}}))\big) \leq t\Big] \cdot \mathbb{P}_{\mathtt{CLIP}}(\mathcal{A}_n(X_{\mathtt{test}})),
\end{equation}

where $\mathcal{H}(p) = -\sum_{i=1}^C p_i \log p_i$ denotes the entropy. The prompt parameters are optimized to minimize the entropy of $\mathbb{P}_{\mathtt{TPT}}(X_{\mathtt{test}})$, thereby improving the reliability of predictions under distributional shifts.



\noindent \textbf{Cache Model.} 
Tip-Adapter~\cite{zhang2022tip} proposes a cache-based strategy to augment CLIP using few-shot samples without requiring retraining. A cache is constructed from $k$-shot features $\mathbf{F}_{\mathtt{train}} = \mathcal{E}_v(x_k)$ and their corresponding one-hot labels $\mathbf{L}_{\mathtt{train}}$ for $N$ classes. During inference, the cache prediction for a test image embedding $f_v = \mathcal{E}_v(X_{\mathtt{test}})$ is computed as:

\begin{equation}
\mathbb{P}_{\mathtt{cache}}(f_v) = A(f_v \mathbf{F}_{\mathtt{train}}^{\top}) \mathbf{L}_{\mathtt{train}},
\end{equation}

where $A(z) = \alpha \exp\left(-\beta(1-z)\right)$ is an adaptation function parameterized by hyperparameters $\alpha$ and $\beta$. The final prediction is obtained by fusing the cache output with the original CLIP head:

\begin{equation}
\mathbb{P}_{\mathtt{TA}}(f_v) = \mathbb{P}_{\mathtt{cache}}(f_v) + f_v \mathbf{W}_c^{\top},
\end{equation}

where $\mathbf{W}_c$ denotes the zero-shot classifier weights from CLIP. This fusion enables the model to leverage both cache-based adaptation and global semantic knowledge for improved generalization.





\subsection{Class-Aware Prototype Cache}

To mitigate cache imbalance and prototype forgetting induced by long-tailed distributions, a \textit{Class-Aware Prototype Cache} (CAPC) module is proposed. The \textit{CAPC} adaptively regulates the cache capacity for each class according to dynamic test-time statistics, thereby enhancing representational diversity and fairness. Two principal mechanisms are integrated within this module: frequency-aware capacity scaling and inactivity-aware rejuvenation.

Let $N_c$ denote the number of cached samples for class $c$, and $N_{\text{total}} = \sum_{i=1}^{C} N_i$ represent the total number of cached samples across all classes. The relative activation frequency for class $c$ is defined as $p_c = \frac{N_c}{N_{\text{total}}}$.

To suppress the dominance of frequent classes and promote equitable cache allocation, a non-linear suppression function is employed:

\begin{equation} 
\phi(p_c) = \tanh\left( -\frac{\log(p_c + \varepsilon)}{s} \right),
\end{equation} 

where $\varepsilon$ prevents divergence for small $p_c$, and $s$ modulates the smoothness of the suppression. This formulation ensures that the adjustment is more pronounced for dominant classes and more accommodating for minority classes. The base cache capacity for class $c$ is determined as: 

\begin{equation} 
M_c = \min\left(M_{\text{max}}, \max\left(1, \left\lceil M \cdot (1 + \gamma \cdot \phi(p_c)) \right\rceil \right)\right), 
\end{equation} 

where $M$ denotes the base capacity, $\gamma$ controls sensitivity to frequency deviations, and $M_{\text{max}}$ sets the upper bound for cache size. Through this adaptive allocation, cache resources are concentrated on rare yet informative classes, thereby improving the model's ability to generalize under imbalanced conditions.

To further alleviate prototype degradation in long-inactive classes, a decay-based rejuvenation strategy is incorporated. Let $t$ be the current iteration and $t_c$ the most recent update timestamp for class $c$. A class is considered inactive if:

\begin{equation} 
t - t_c > \eta, 
\end{equation} 

where $\eta$ specifies the inactivity threshold. For such classes, a frequency-decayed capacity boost is computed as follows:

\begin{equation} 
M_c^{\text{boost}} = \left\lceil \delta \cdot e^{-\alpha p_c} \cdot \frac{t - t_c}{\eta} \right\rceil,
\end{equation} 

where $\delta$ sets the maximum boost magnitude and $\alpha$ penalizes frequent classes to maintain fair cache distribution. The final cache capacity for class $c$ is given by:

\begin{equation} 
M_c^{\text{total}} = M_c + M_c^{\text{boost}}, 
\end{equation} 

This decay-aware compensation mechanism enables inactive or tail classes to preserve robust prototype representations, even under highly skewed adaptation scenarios. By dynamically balancing cache allocation and rejuvenation, the \textit{CAPC} module substantially mitigates the adverse effects of class imbalance and prototype forgetting during test-time adaptation.

\subsection{Negative Contrastive Learning}

Class confusion among semantically similar categories remains a significant obstacle in test-time adaptation, often leading to misclassification and reduced generalization. To address this issue, the \textit{Negative Contrastive Learning} (NCL) mechanism is introduced, wherein cross-modal prototypes exhibiting maximal interference are actively selected and contrastive alignment constraints are imposed to enhance inter-class separability.

For each class $c$, let $v_c$ and $t_c$ denote its visual and textual prototypes, respectively. The most similar negative prototypes from all other classes are identified as follows:

\begin{equation}
v_c^{-} = \arg\max_{j \ne c} \cos(v_c, v_j), \quad
t_c^{-} = \arg\max_{j \ne c} \cos(t_c, t_j),
\end{equation}

By focusing on these hard negatives, a robust InfoNCE loss~\cite{oord2018representation} is constructed for class $c$:

\begin{equation}
\mathcal{L}_{\text{NCL}}(c) = - \log \left( \frac{ \exp(\cos(v_c, t_c)/\tau) }{ \exp\left(\frac{\cos(v_c, t_c)}{\tau}\right) + \exp\left(\frac{\cos(v_c, t_c^{-})}{\tau}\right) + \exp\left(\frac{\cos(v_c^{-}, t_c)}{\tau}\right) } \right),
\end{equation}
where $\tau$ denotes the temperature parameter controlling the sharpness of the contrastive distribution. This formulation enforces explicit separation between the prototypes of semantically similar classes by penalizing close associations in both visual and textual modalities.

The aggregate hard negative contrastive loss is computed as the average over all currently activated classes:
\begin{equation}
\mathcal{L}_{\text{NCL}} = \frac{1}{|\mathcal{C}|} \sum_{c \in \mathcal{C}} \mathcal{L}_{\text{NCL}}(c),
\end{equation}
where $\mathcal{C}$ represents the set of classes currently maintained in the cache. By integrating this mechanism, the model is encouraged to maintain fine-grained discriminability at class boundaries, thereby improving robustness and generalization in open-world and long-tailed adaptation scenarios. The \textit{NCL} module thus serves as a crucial component for mitigating semantic overlap and enhancing the reliability of test-time predictions.

\subsection{Loss and Optimization}
In contrast to DPE~\cite{zhang2024dual}, which symmetrically updates both visual and textual prototypes, our approach adopts a sophisticated asymmetric adaptation strategy. Visual prototypes are dynamically maintained through the \textit{CAPC} mechanism, which adaptively updates cache contents based on prediction confidence and class frequency statistics, while textual prototypes undergo direct parametric refinement during adaptation. This design leverages the stability of memory-based visual representations while enabling targeted cross-modal alignment through adaptive textual prototype learning.

The overall training objective integrates three key components:
\begin{equation}
\mathcal{L}_{\text{total}} = \mathcal{L}_{\text{aug}} + \lambda_1 \mathcal{L}_{\text{align}} + \lambda_2 \mathcal{L}_{\text{NCL}},
\end{equation}
where $\mathcal{L}_{\text{aug}}$ encourages confident predictions by minimizing entropy, $\mathcal{L}_{\text{align}}$ enforces cross-modal alignment between textual and visual prototypes via InfoNCE, and $\mathcal{L}_{\text{NCL}}$ promotes inter-class separation using the Negative Contrastive Learning mechanism. The hyperparameters $\lambda_1$ and $\lambda_2$ modulate the relative contribution of each auxiliary term, allowing for flexible balancing of adaptation objectives.

During inference, the final class probability for a given visual feature $f_v$ is computed by fusing the refined textual prototype $\hat{t}_c$ with the cached visual prototype $v_c$:

\begin{equation}
\mathbb{P}(y = c \mid X) = \frac{\exp\left( \frac{f_v^\top \hat{t}_c + \alpha \exp(-\beta (1 - f_v^\top v_c))}{\tau} \right)}
{\sum_{j=1}^{C} \exp\left( \frac{f_v^\top \hat{t}_j + \alpha \exp(-\beta (1 - f_v^\top v_j))}{\tau} \right)},
\end{equation}
where $\alpha$ and $\beta$ are fusion parameters controlling the influence of memory stability, and $\tau$ denotes the temperature for normalization.

Through this asymmetric optimization and fusion framework, the model achieves robust generalization across diverse distribution shifts. The \textit{CAPC} dynamically maintains visual prototypes with frequency-aware capacity scaling and inactivity-based rejuvenation, ensuring resilient memory adaptation, while parametric textual prototype refinement enables efficient cross-modal alignment. This complementary adaptation strategy balances memory robustness with cross-modal flexibility for effective test-time generalization.

\section{Experiments}

\subsection{Benchmarks.}
The proposed methods are evaluated under two complementary test-time adaptation scenarios: robustness to distribution shifts and generalization across heterogeneous domains. To rigorously assess robustness, the out-of-distribution (OOD) benchmark is adopted, wherein models are exposed to samples that deviate significantly from the training distribution. Four challenging datasets derived from ImageNet~\cite{deng2009imagenet} are utilized for this purpose: ImageNet-A~\cite{hendrycks2021natural}, ImageNet-V2~\cite{recht2019imagenet}, ImageNet-R~\cite{hendrycks2021many}, and ImageNet-S~\cite{wang2019learning}. Each dataset introduces distinct forms of perturbations and distributional shifts, thereby providing a stringent evaluation of the model's ability to maintain performance under unforeseen conditions. To further examine cross-domain generalization, a comprehensive benchmark comprising ten diverse image classification datasets is employed. These datasets span a wide array of visual domains, including Aircraft~\cite{maji2013fine}, Caltech101~\cite{fei2004learning}, Cars~\cite{krause20133d}, DTD~\cite{cimpoi2014describing}, EuroSAT~\cite{helber2019eurosat}, Flower102~\cite{nilsback2008automated}, Food101~\cite{bossard2014food}, Pets~\cite{parkhi2012cats}, SUN397~\cite{xiao2010sun}, and UCF101~\cite{soomro2012ucf101}. Unlike the OOD benchmark, this setting requires models to transfer knowledge across domains with disjoint class spaces, thereby highlighting their adaptability to novel visual environments. By systematically evaluating under both OOD and cross-domain benchmarks, a comprehensive understanding of the model's performance is obtained. This dual perspective enables robust measurement of both resilience to unseen distributional shifts and flexibility in adapting to a wide range of visual domains, thus providing a holistic assessment of test-time adaptation capabilities.

\begin{table*}[t]
\renewcommand\arraystretch{1}
\renewcommand{\tabcolsep}{2pt}
\caption{Evaluation of model robustness under natural distribution shifts across five OOD datasets, using CLIP’s ResNet-50 and ViT-B/16 backbones. We report top-1 accuracy (\%) across all methods. \textbf{Bold} and \underline{Underlined} values indicate the best and second-best results, respectively.}
\vspace{0pt}
\centering
\resizebox{1.00\linewidth}{!}{
\begin{tabular}{lr||cccccc|c}
\toprule
Method & Source & ImageNet & ImageNet-A & ImageNet-V2 & ImageNet-R & ImageNet-S & {Average} & {OOD Average} \\
\midrule
CLIP-ResNet-50~\cite{radford2021learning} & ICML'21 & 58.16 & 21.83 & 51.41 & 56.15 & 33.37 & 44.18 & 40.69 \\ 
\midrule
CoOp~\cite{zhou2022learning} & IJCV'22 & 63.33 & 23.06 & 55.40 & 56.60 & 34.67 & 46.61 & 42.43 \\
CoCoOp~\cite{zhou2022conditional} & CVPR'22 & 62.81 & 23.32 & 55.72 & 57.74 & 34.48 & 46.81 & 42.82 \\
Tip-Adapter~\cite{zhang2022tip} & ECCV'22 & 62.03 & 23.13 & 53.97 & 60.35 & 35.74 & 47.04 & 43.30 \\

\midrule
TPT~\cite{shu2022test} & NeurIPS'22 & 60.74 & 26.67 & 54.70 & 59.11 & 35.09 & 47.26 & 43.89 \\
DiffTPT~\cite{feng2023diverse}& ICCV'23 & 60.80 & \underline{31.06} & 55.80 & 58.80 & 37.10 & 48.71 & 45.69 \\
TDA~\cite{karmanov2024efficient} & CVPR'24 & 61.35 & 30.29 & 55.54 & 62.58 & 38.12 & 49.58 & 46.63 \\ 
DMN-ZS~\cite{zhang2024dual0} & CVPR'24 & \underline{63.87} & 28.57 & 56.12 & 61.44 & 39.84 & 49.97 & 46.49 \\ 
DPE~\cite{zhang2024dual} & NeurIPS'24 & 63.41 & 30.15 &  \underline{56.72} &  \underline{63.72} & \underline{40.03} & \underline{50.81} & \underline{47.66} \\  

TPS~\cite{sui2025just} & WACV'25 & 61.47 & 30.48 & 54.96 & 62.87 & 37.14 & 49.38 & 46.36 \\

\rowcolor{gray!20}

\textbf{CPL-NC} & \textbf{Ours} & \textbf{64.13} & \textbf{31.09} & \textbf{57.48} & \textbf{64.25} & \textbf{40.64} & \textbf{51.52} & \textbf{48.37}\\

\midrule
\midrule
CLIP-ViT-B/16~\cite{radford2021learning} & ICML'21 & 66.73 & 47.87 & 60.86 & 73.98 & 46.09 & 59.11 & 57.20 \\
\midrule
CoOp~\cite{zhou2022learning} & IJCV'22 & 71.51 & 49.71 & 64.20 & 75.21 & 47.99 & 61.72 & 59.28 \\
CoCoOp~\cite{zhou2022conditional} & CVPR'22 & 71.02 & 50.63 & 64.07 & 76.18 & 48.75 & 62.13 & 59.91 \\
Tip-Adapter~\cite{zhang2022tip} & ECCV'22 & 70.75 & 51.04 & 63.41 & 77.76 & 48.88 & 62.37 & 60.27 \\
\midrule
TPT~\cite{shu2022test} & NeurIPS'22 & 68.98 & 54.77 & 63.45 & 77.06 & 47.94 & 62.44 & 60.81 \\
DiffTPT~\cite{feng2023diverse} & ICCV'23 & 70.30 & 55.68 & 65.10 & 75.00 & 46.80 & 62.28 & 60.52 \\
TDA~\cite{karmanov2024efficient} & CVPR'24 & 69.51 & 60.11 & 64.67 & 80.24 & 50.54 & 65.01 & 63.89 \\ 

DMN-ZS~\cite{zhang2024dual0} & CVPR'24 & \textbf{72.25} & 58.28 & 65.17 & 78.55 & \underline{53.20} & 65.49 & 63.80 \\
DPE~\cite{zhang2024dual} & NeurIPS'24 & 71.91 & 59.63 &  \underline{65.44} &  \underline{80.40} & 52.26 &  \underline{65.93} &  \underline{64.43} \\ 
TTL~\cite{imam2025test} & WACV'25 & 70.23 & \textbf{60.51} & 64.55 & 77.54 & 48.61 & 64.29 & 62.80 \\
TPS~\cite{sui2025just}& WACV'25 & 70.19 & 60.08 & 64.73 & 80.27 & 49.95 & 65.04 & 63.76 \\

\rowcolor{gray!20}
\textbf{CPL-NC} & \textbf{Ours} & \underline{72.17} & \underline{60.31} & \textbf{65.95} & \textbf{80.95} & \textbf{53.53} & \textbf{66.58} & \textbf{65.19} \\

\bottomrule
\end{tabular}
}

\label{tab:ood-main}
\vspace{-10pt}
\end{table*}

\subsection{Implementation Details.}
All models are initialized from pre-trained CLIP~\cite{radford2021learning} backbones, specifically ResNet-50 and ViT-B/16 architectures. Test-time adaptation is conducted on individual images, with a batch size fixed at 1, and no manual tuning of hand-crafted prompts is performed. The cache mechanism is designed to update features selectively based on prediction entropy, giving priority to samples with lower entropy to enhance the reliability of stored representations. Furthermore, cache capacities are dynamically managed in a class-aware manner, being adjusted according to sample frequency via a hyperparameter $\gamma$, and each class cache is capped at a maximum of 10 entries to maintain computational efficiency. Adaptation is achieved by jointly optimizing three objectives: entropy minimization, visual-textual prototype alignment, and negative contrastive loss. Negative prototypes are updated periodically to ensure effective separation between confusing classes. The AdamW optimizer is utilized for all adaptation steps, with learning rates specified in configuration files to ensure reproducibility. Notably, no source training data is accessed during test-time adaptation, preserving the integrity of the evaluation protocol. All experiments are executed on a single NVIDIA RTX 4090 GPU. Performance is consistently reported using top-1 accuracy (\%), enabling fair comparison across different settings and benchmarks. This implementation protocol ensures both efficiency and rigor in the evaluation of test-time adaptation methods.

\subsection{Results and Discussions} 

\subsubsection{Evaluation under Natural Out-of-Distribution Shifts.} 

Robustness under natural distribution shifts is systematically evaluated across five OOD datasets: ImageNet-A~\cite{hendrycks2021natural}, ImageNet-V2~\cite{recht2019imagenet}, ImageNet-R~\cite{hendrycks2021many}, ImageNet-S~\cite{wang2019learning}, and the original ImageNet~\cite{deng2009imagenet}, all under the unsupervised test-time adaptation setting. As reported in Table~\ref{tab:ood-main}, CPL-NC, with ResNet-50 as the visual encoder, consistently achieves superior performance on all five benchmarks. Notably, CPL-NC attains 31.09\% Top-1 accuracy on the highly challenging ImageNet-A~\cite{hendrycks2021natural} and 57.48\% on ImageNet-V2~\cite{recht2019imagenet}, surpassing the previous state-of-the-art DPE~\cite{zhang2024dual}. On average, CPL-NC yields 51.52\%, outperforming DPE by +0.71\%, which indicates enhanced overall robustness. When ViT-B/16 serves as the backbone, CPL-NC maintains its advantage, achieving 65.95\% on ImageNet-V2 and 80.95\% on ImageNet-R, both exceeding the corresponding results from DPE. The average performance reaches 66.58\%, maintaining the top performance.

The observed performance improvements can be drawn from the architectural design of CPL-NC. The \textit{CAPC} module and \textit{NCL} mechanism are specifically constructed to address key challenges in OOD adaptation. The CAPC module enhances memory stability by dynamically allocating cache capacity according to class frequency and employing a decay-aware strategy to revive rare-class prototypes. This approach mitigates the risk of prototype collapse and ensures robust representation for under-represented classes, as reflected by the observed gains over TDA~\cite{karmanov2024efficient} and DPE~\cite{zhang2024dual} on ImageNet-A. The NCL mechanism leverages hard cross-modal negatives to enforce triplet-based constraints, thereby sharpening semantic boundaries and improving discriminative power. This effect is particularly pronounced on datasets such as ImageNet-R and ImageNet-S, which contain semantically similar categories (e.g., “clownfish” vs. “goldfish”, “windmill” vs. “water tower”), where CPL-NC demonstrates clear improvements over previous methods.

Theoretically, these results suggest that the complementary interaction between CAPC and NCL modules not only stabilizes memory but also enhances semantic discrimination during adaptation. By addressing both prototype reliability and boundary refinement, CPL-NC is enabled to generalize more effectively under distributional shifts. The consistent outperformance across diverse backbones and datasets substantiates the robustness and adaptability of the proposed approach.

\begin{table*}[t]
\renewcommand\arraystretch{1.15}
\renewcommand{\tabcolsep}{2pt}
\caption{Cross-Dataset Generalization Performance. We report top-1 accuracy (\%) of various methods on 10 diverse target datasets using CLIP~\cite{radford2021learning} backbones (ResNet-50 and ViT-B/16), evaluating their ability to generalize to unseen domains. \textbf{Bold} and \underline{Underlined} values denote the best and second-best results, respectively.}
  \vspace{0pt}
  \centering
  \resizebox{1.00\linewidth}{!}{
    \begin{tabular}{lr||cccccccccc|c}
      \toprule
      Method & Source  & Aircraft & Caltech & Cars & DTD & EuroSAT & Flower & Food101 & Pets & SUN397 & UCF101 & Average \\
      \midrule
      CLIP-ResNet50~\cite{radford2021learning}& ICML'21 & 15.66 & 85.88 & 55.70 & 40.37 & 23.69 & 61.75 & 73.97 & 83.57 & 58.80 & 58.84 & 55.82 \\
      \midrule
      CoOp~\cite{zhou2022learning} & IJCV'22 & 15.12 & 86.53 & 55.32 & 37.29 & 26.20 & 61.55 & 75.59 & 87.00 & 58.15 & 59.05 & 56.18 \\
      CoCoOp~\cite{zhou2022conditional} & CVPR'22 & 14.61 & 87.38 & 56.22 & 38.53 & 28.73 & 65.57 & 76.20 & 88.39 & 59.61 & 57.10 & 57.23 \\
      Tip-Adapter~\cite{zhang2022tip} & ECCV'22 & 16.11 & 87.26 & 55.89 & 40.37 & 25.79 & 62.77 & 74.82 & 82.97 & 60.85 & 59.48 & 56.63 \\
      
      \midrule
      TPT~\cite{shu2022test} & NeurIPS'22 & 17.58 & 87.02 & 58.46 & 40.84 & 28.33 & 62.69 & 74.88 & 84.49 & 61.46 & 60.82 & 57.66 \\
      DiffTPT~\cite{feng2023diverse} & ICCV'23 & 17.60 & 86.89 &  \textbf{60.71} & 40.72 & 41.04 & 63.53 &\textbf{79.21} & 83.40 & 62.72 & 62.67 & 59.85 \\
      TDA~\cite{karmanov2024efficient} & CVPR'24 & 17.61 & 89.70 & 57.78 & 43.74 &  \underline{42.11} &  \underline{68.74} & 77.75 &  86.18 & 62.53 &  \textbf{64.18}  & 61.03\\ 
      DPE~\cite{zhang2024dual} & NeurIPS'24 &  \underline{19.80} &  \underline{90.83} & 59.26 &  \underline{50.18} & 41.67 & 67.60 & 77.83 & 85.97 & \underline{64.23} & 61.98 & \underline{61.93} \\ 
      PCPT~\cite{wang2025ctpt} & PR'25 &  17.10 &  87.20 & 54.40 & 43.20 & 36.90 & 65.90 & 75.60 & \textbf{87.90} & 60.80 & \underline{63.00} & 59.10 \\  \rowcolor{gray!20}
      \textbf{CPL-NC} & \textbf{Ours} & \textbf{22.23} & \textbf{91.28} & \underline{59.91} & \textbf{53.37} & \textbf{45.67} & \textbf{69.67} & \underline{78.56} & \underline{86.29} & \textbf{64.64} & 63.05 & \textbf{63.47} \\
      \midrule
      \midrule
      CLIP-ViT-B/16~\cite{radford2021learning} & ICML'21 & 23.67 & 93.35 & 65.48 & 44.27 & 42.01 & 67.44 & 83.65 & 88.25 & 62.59 & 65.13 & 63.58 \\
      \midrule
      CoOp~\cite{zhou2022learning} & IJCV'22 & 18.47 & 93.70 & 64.51 & 41.92 & 46.39 & 68.71 & 85.30 & 89.14 & 64.15 & 66.55 & 63.88 \\
      CoCoOp~\cite{zhou2022conditional} & CVPR'22  & 22.29 & 93.79 & 64.90 & 45.45 & 39.23 & 70.85 & 83.97 & 90.46 & 66.89 & 68.44 & 64.63 \\
      Tip-Adapter~\cite{zhang2022tip} & ECCV'22 & 16.11 & 87.26 & 55.89 & 40.37 & 25.79 & 62.77 & 74.82 & 82.97 & 60.85 & 59.48 & 56.63 \\      
      \midrule
      TPT~\cite{shu2022test} & NeurIPS'22 & 24.78 & 94.16 & 66.87 & 47.75 & 42.44 & 68.98 & 84.67 & 87.79 & 65.50 & 68.04 & 65.10 \\
      DiffTPT~\cite{feng2023diverse} & ICCV'23 & 25.60 & 92.49 & 67.01 & 47.00 & 43.13 & 70.10 & \textbf{87.23} & 88.22 & 65.74 & 62.67 &65.47 \\
      TDA~\cite{karmanov2024efficient} & CVPR'24 & 23.91 & 94.24 & 67.28 
      & 47.40 & \underline{58.00} & 71.42 & 86.14 & 88.63 & 67.62 & 70.66 & 67.53 \\ 
      DPE~\cite{zhang2024dual} & NeurIPS'24 & \underline{28.95} & \underline{94.81} & 67.31 & \underline{54.20} & 55.79 & \underline{75.07} & 86.17 & 91.14 & \textbf{70.07} & 70.44 & \underline{69.40} \\
      TTL~\cite{imam2025test}  & WACV'25 & 23.82 & 93.63 & \underline{67.96} & 46.69 & 42.02 & 70.48 & 85.05 & 88.72 & 66.32 & 69.20 & 65.39  \\ 
      PCPT~\cite{wang2025ctpt} & PR'25 &  26.00 &  94.10 & 64.70 & 47.20 & 48.80 & 73.30 & 84.60 & \textbf{92.20} & 65.60 & \textbf{71.10} & 66.80 \\  
      \rowcolor{gray!20}
      \textbf{CPL-NC} & \textbf{Ours} & \textbf{29.82} & \textbf{95.50} & \textbf{68.62} & \textbf{55.67} & \textbf{58.15} & \textbf{77.18} & \underline{86.37} & \underline{91.55} & \underline{70.05} & \underline{70.71} & \textbf{70.36 }\\

      \bottomrule
    \end{tabular}
  } 

  \label{tab:fine-grained}
  \vspace{-5pt}
\end{table*}

\subsubsection{Generalization Across Diverse Datasets.} 

To assess cross-dataset generalization, each method is evaluated on ten heterogeneous target domains without any target-specific fine-tuning. As summarized in Table~\ref{tab:fine-grained}, CPL-NC consistently demonstrates superior transferability under the CLIP-ResNet50 backbone, achieving the highest top-1 accuracy on 6 out of 10 datasets and yielding an average of 63.47\%. This result surpasses the previous state-of-the-art DPE~\cite{zhang2024dual} by +1.54\%. Notably, CPL-NC achieves 22.23\% accuracy on the fine-grained Aircraft dataset~\cite{maji2013fine} and 53.37\% on the texture-centric DTD dataset~\cite{cimpoi2014describing}, exceeding DPE by +2.43\% and +3.19\%, respectively. These improvements highlight the method’s robustness to structural and local pattern shifts. When evaluated with the ViT-B/16 backbone, CPL-NC further extends its advantage, leading on 9 out of 10 datasets and achieving an average accuracy of 70.36\%. This performance exceeds both DPE~\cite{zhang2024dual} (69.40\%) and TDA~\cite{karmanov2024efficient} (67.53\%). For instance, CPL-NC attains 58.15\% on EuroSAT~\cite{helber2019eurosat}, a remote sensing domain, and 70.71\% on the action recognition benchmark UCF101~\cite{soomro2012ucf101}, both outperforming DPE.

The observed generalization gains can be theoretically attributed to the adaptive mechanisms embedded within CPL-NC. The dynamic cache allocation and negative contrastive learning strategies are designed to enhance prototype reliability and refine semantic boundaries, which are essential for transferring knowledge across domains with varying visual and semantic characteristics. By mitigating the effects of domain-specific biases and promoting robust feature alignment, CPL-NC is enabled to maintain high accuracy even under substantial distributional and pattern shifts. These consistent improvements across diverse datasets and backbones substantiate the method’s effectiveness in generalizing beyond the source domain. The results provide strong evidence for the theoretical soundness and broad applicability of CPL-NC in handling diverse semantic and visual distributions.

\subsubsection{Efficiency Comparison.} 

The inference efficiency and adaptation performance of mainstream test-time adaptation (TTA) methods are systematically evaluated on ImageNet~\cite{deng2009imagenet}, with results presented in Table~\ref{table:efficiency}. Compared to the zero-shot CLIP~\cite{radford2021learning} baseline, which achieves 59.81\% top-1 accuracy in 11 minutes, CPL-NC attains a substantially higher accuracy of 64.13\% after 2 hours and 44 minutes of adaptation, yielding a notable improvement of +4.32\%. Although TPT~\cite{shu2022test} and DiffTPT~\cite{feng2023diverse} reach comparable accuracies of 60.74\% and 60.80\%, these methods require more than 10 and 20 hours of computation, respectively. Such extended runtimes impose significant computational overhead, limiting their practical utility in scenarios where efficiency is essential. TDA~\cite{karmanov2024efficient} demonstrates greater efficiency with a runtime of 32 minutes, yet this is accompanied by a reduced accuracy of 61.35\%. These results illustrate that CPL-NC achieves the highest accuracy among all evaluated methods while maintaining a moderate adaptation time. The adaptive mechanisms in CPL-NC, such as dynamic cache allocation and negative contrastive learning, are considered to facilitate rapid convergence and robust performance. It can be observed that CPL-NC provides a more effective balance between adaptation speed and accuracy compared to alternative approaches, further highlighting its methodological advantages in test-time adaptation settings.

\begin{table*}[t]
  \centering
  \caption{Efficiency and accuracy comparison on ImageNet~\cite{deng2009imagenet}. We report test-time adaptation duration, final top-1 accuracy, and accuracy improvement over zero-shot CLIP~\cite{radford2021learning}.}
  \small
  \begin{tabular}{l||ccc}
    \toprule
    Method  & Time & Acc & Gain \\ 
    \midrule
    CLIP~\cite{radford2021learning}  & 11 min  &  59.81&  -\\
    TPT~\cite{shu2022test} &   $>$10 h &  60.74  & +0.93  \\
    DiffTPT~\cite{feng2023diverse} &  $>$20 h &  60.80  & +0.99  \\
    TDA~\cite{karmanov2024efficient} & 32 min  &  61.35  & +1.54  \\
    DPE~\cite{zhang2024dual} & 2 h 37 min & 63.41 & +3.60 \\
    \rowcolor{gray!20}
    \textbf{CPL-NC (ours)} & 2 h 44 min & \textbf{64.13} & \textbf{+4.32} \\
    \bottomrule
  \end{tabular}
  \label{table:efficiency}
  \vspace{-10pt}
\end{table*}

\section{Ablation Study}

\begin{table*}[t]
\centering
\caption{Ablation study of CAPC and NCL modules on ImageNet, ImageNet-A, and Aircraft datasets.}
\label{tab:ablation1}
\renewcommand\arraystretch{1.0}
\renewcommand{\tabcolsep}{8pt}
\small
\resizebox{0.75\linewidth}{!}{
\begin{tabular}{ccc||ccc}
\toprule
\# & CAPC & NCL & ImageNet Acc. & ImageNet-A Acc. & Aircraft Acc. \\
\midrule
1 & \ding{55} & \ding{55} & 63.12 & 59.47 & 19.70 \\
2 & \ding{55} & \ding{51} & 63.33 & 59.79 & 20.76 \\
3 & \ding{51} & \ding{55} & 63.75 & 59.73 & 21.08 \\
\rowcolor{gray!20}
4 & \ding{51} & \ding{51} & \textbf{64.13} & \textbf{60.31} & \textbf{22.23} \\
\bottomrule
\end{tabular}
}
\vspace{-10pt}
\end{table*}

\subsubsection{Analyzing the Contributions of CAPC and NCL.} 

Ablation studies were conducted on ImageNet~\cite{deng2009imagenet}, ImageNet-A~\cite{hendrycks2021natural}, and Aircraft~\cite{maji2013fine} to evaluate the impact of the CAPC and NCL modules. As reported in Table~\ref{tab:ablation1}, the baseline model, which excludes both CAPC and NCL, achieves an accuracy of 63.12\% on ImageNet, 59.47\% on ImageNet-A, and 19.70\% on Aircraft. When only the NCL module is introduced, the accuracy increases to 63.33\% on ImageNet, 59.79\% on ImageNet-A, and 20.76\% on Aircraft. This improvement is more evident on the Aircraft dataset, which contains many visually similar categories, indicating that NCL effectively alleviates confusion among fine-grained classes and enhances semantic discrimination. The introduction of CAPC alone results in a further increase in accuracy, reaching 63.75\% on ImageNet, 59.73\% on ImageNet-A, and 21.08\% on Aircraft. The gains brought by CAPC are particularly notable on the Aircraft dataset, suggesting that its ability to dynamically regulate cache capacity and preserve the representations of rare classes contributes significantly to overall performance. When both CAPC and NCL modules are employed together, the model achieves the highest accuracy across all datasets, with 64.13\% on ImageNet, 60.31\% on ImageNet-A, and 22.23\% on Aircraft. This result demonstrates that the two modules complement each other: CAPC strengthens structural memory and ensures balanced representation for all classes, while NCL reduces semantic ambiguity and improves discrimination between closely related categories. The combined effect of these modules leads to robust generalization, especially under distribution shifts and in settings characterized by high semantic complexity. Overall, the experimental results confirm that both CAPC and NCL are indispensable for optimal adaptation and that their joint deployment substantially enhances the effectiveness of the CPL-NC framework.

\begin{figure*}[t]
  \centering
  \includegraphics[width=1.00\textwidth]{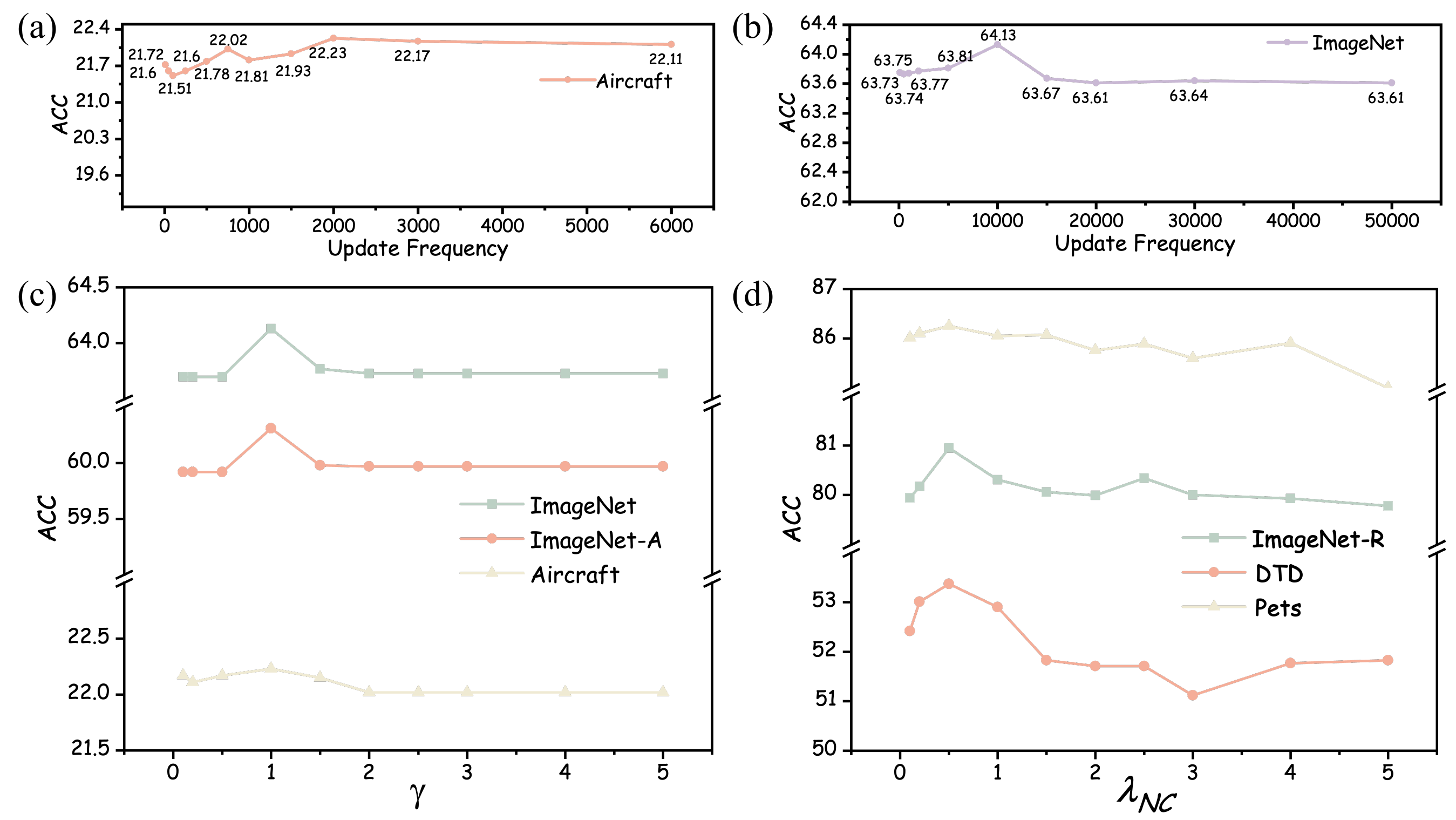}
  \caption{(a) Impact of update frequency on model performance evaluated on Aircraft, and (b) ImageNet; (c) influence of the class-aware factor $\gamma$ on ImageNet, ImageNet-A, and Aircraft; (d) effect of the NCL loss weight on model generalization across ImageNet-R, DTD, and Pets datasets.}
  \label{ablation}
  \vspace{-10pt}
\end{figure*}

\subsubsection{Hyperparameter Sensitivity Analysis.} 

To systematically evaluate the influence of key hyperparameters in CPL-NC, a series of controlled experiments were conducted across multiple datasets, with the results presented in Figure~\ref{ablation}. The impact of update frequency was first examined. As depicted in Figure~\ref{ablation}(a) and (b), accuracy on Aircraft and ImageNet datasets peaked when the update frequency was set to approximately 20\% of the total test samples. Increasing the update frequency beyond this threshold led to reduced accuracy, likely due to excessive noise and instability, while lower frequencies delayed adaptation and resulted in suboptimal performance. These findings indicate that an intermediate update frequency is crucial for maintaining efficient and stable adaptation. The cache scaling factor $\gamma$ in the CAPC module was also analyzed. Figure~\ref{ablation}(c) shows that setting $\gamma = 1.0$ consistently resulted in strong performance across ImageNet, ImageNet-A, and Aircraft. Both higher and lower values of $\gamma$ failed to achieve optimal results, suggesting that only moderate scaling of class frequency effectively balances cache representation without causing capacity allocation issues. Sensitivity to the NCL loss weight $\lambda_{\text{NC}}$ was further investigated. As shown in Figure~\ref{ablation}(d), the choice of $\lambda_{\text{NC}} = 0.5$ produced the highest accuracy on ImageNet-R, DTD, and Pets. Excessively large values introduced gradient conflicts and reduced class separability, whereas smaller values weakened the discriminative power of the NCL module. This demonstrates the necessity of moderate loss weighting to enhance semantic discrimination while preserving training stability. Taken together, these analyses demonstrate that CPL-NC is robust to hyperparameter variations within reasonable ranges, and that careful adjustment of update frequency, cache scaling, and loss weighting is essential for optimal test-time adaptation.

\section{Conclusion}

We propose CPL-NC, a test-time adaptation framework tailored for vision-language models, addressing two critical challenges in real-world scenarios: class imbalance and semantic confusion. To this end, CPL-NC introduces a CAPC Module, which enhances representation stability for tail classes via frequency-driven capacity allocation and a rejuvenation strategy based on activation recency. In parallel, a Negative Contrastive Learning mechanism imposes contrastive constraints between the most semantically confusing cross-modal pairs, effectively improving class separability. The overall framework adopts an asymmetric optimization strategy that updates only textual prototypes, ensuring both inference efficiency and representation reliability. Extensive experiments on 15 cross-domain datasets demonstrate that CPL-NC significantly enhances robustness and generalization without requiring target-domain labels.

\bibliography{reference}

\end{document}